\documentclass[pmlr,twocolumn,10pt]{jmlr} %

\usepackage{booktabs}
\usepackage{siunitx}

\definecolor{todo-color}{HTML}{ff3232}

\definecolor{green}{cmyk}{1,0,1,0}

\definecolor{amethyst}{rgb}{0.6, 0.4, 0.8}

\usepackage[switch]{lineno}

\theorembodyfont{\upshape}
\theoremheaderfont{\scshape}
\theorempostheader{:}
\theoremsep{\newline}

\jmlrvolume{LEAVE UNSET}
\jmlryear{2025}
\jmlrsubmitted{LEAVE UNSET}
\jmlrpublished{LEAVE UNSET}
\jmlrworkshop{Conference on Health, Inference, and Learning (CHIL) 2025} %

\title[Revealing Treatment Non‑Adherence Bias in Clinical Machine Learning Using Large Language Models]{Revealing Treatment Non‑Adherence Bias in Clinical Machine Learning Using Large Language Models}

\author{%
\Name{Zhongyuan Liang} \Email{zhongyuan\_liang@berkeley.edu}\\
\addr UC Berkeley and UCSF
\AND
\Name{Arvind Suresh} \Email{arvind.suresh@ucsf.edu}\\
\addr UCSF
\AND
\Name{Irene Y. Chen} \Email{iychen@berkeley.edu}\\
\addr UC Berkeley and UCSF
}

\begin{document}

\maketitle

\begin{abstract}
Machine learning systems trained on electronic health records (EHRs) increasingly guide treatment decisions, but their reliability depends on the critical assumption that patients follow the prescribed treatments recorded in EHRs.
Using EHR data from 3,623 hypertension patients, we investigate how treatment non-adherence introduces implicit bias that can fundamentally distort both causal inference and predictive modeling.
By extracting patient adherence information from clinical notes using a large language model (LLM), we identify 786 patients (21.7\%) with medication non-adherence. We further uncover key demographic and clinical factors associated with non-adherence, as well as patient-reported reasons including side effects and difficulties obtaining refills.
Our findings demonstrate that this implicit bias can not only reverse estimated treatment effects, but also degrade model performance by up to 5\% while disproportionately affecting vulnerable populations by exacerbating disparities in decision outcomes and model error rates.
This highlights the importance of accounting for treatment non-adherence in developing responsible and equitable clinical machine learning systems.

\end{abstract}

\paragraph*{Data and Code Availability}
Our study utilizes EHR data from UCSF. Details on the data and preprocessing steps are provided in the following sections. While we will make the code publicly available, the data cannot be shared due to data use agreement.

\paragraph*{Institutional Review Board (IRB)}
Our work was reviewed by an IRB and declared exempt as it focuses on retrospective analysis using de-identified data.

\begin{figure*}[th]
    \centering
    \includegraphics[width=\textwidth]{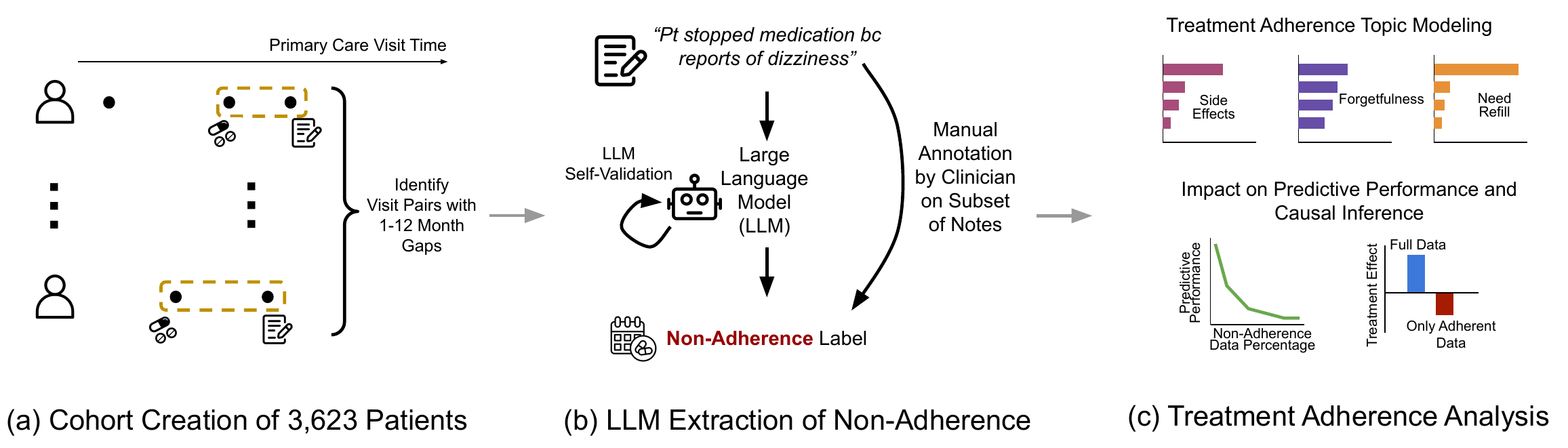}
    \caption{Illustration of cohort selection, LLM non-adherence extraction, and non-adherence analysis. (a) We select 3,623 hypertension patients and pair their visits, with hypertension medication prescribed at the first visit and clinical notes extracted from the second. (b) These notes are then processed by an LLM to identify treatment non-adherence, with outputs validated through clinician annotations. (c) We further perform topic modeling to uncover reasons for non-adherence and assess the harmful impact of ignoring this bias on predictive modeling performance and treatment effect estimation.}
    \label{fig:diagram}
\end{figure*}

\section{Introduction}
\label{sec:intro}
Treatment non-adherence is a pervasive and persistent challenge in healthcare. Researchers estimate that poor medication adherence leads to 125,000 preventable deaths annually in the U.S. and contributes to \$100-\$300 billion in avoidable healthcare costs \citep{benjamin2012medication}. This issue is particularly prevalent among patients with chronic conditions such as hypertension, with 40-50\% failing to take their medications as prescribed \citep{kleinsinger2018unmet, algabbani2020treatment}. While researchers have extensively documented this problem through surveys and interviews \citep{boratas2018evaluation, fernandez2019adherence, algabbani2020treatment, najjuma2020adherence, schober2021high}, the studies---and ultimately understanding of treatment non-adherence---remain limited by small sample sizes and self-reporting bias \citep{adams1999evidence, stirratt2015self}. Physical solutions to monitor and encourage adherence such as electronic pill caps have shown promise in controlled settings but remain impractical for large-scale deployment due to high costs and implementation challenges~\citep{parker2007adherence,mauro2019effect}.
 
These measurement challenges take on new urgency as healthcare systems increasingly rely on machine learning (ML) models trained on electronic health records (EHRs) to guide treatment decisions \citep{komorowski2018artificial, brugnara2020multimodal, zheng2021personalized, mroz2024predicting, yi2024development,shen2024data, chen2022clustering}. These machine learning models learn from historical patient data, which assume that prescribed treatments were actually taken. However, this introduces an implicit bias---models trained on non-adherent patients learn patterns that misrepresent true treatment effects. This implicit bias may degrade model performance and disproportionately impact underserved populations, who often face greater barriers to treatment adherence \citep{bosworth2006racial, schober2021high}.

Recent advances in large language models (LLMs) have shown that LLMs can advance medical understanding by accurately extracting information from EHRs \citep{agrawal2022large, goel2023llmsaccelerateannotationmedical}. Instead of relying on self-reported treatment adherence from questionnaires and interviews, LLMs could serve as a powerful tool for identifying treatment non-adherence directly from EHRs. By analyzing rich but unstructured clinical notes, LLMs can detect documented instances of missed medications, unfilled prescriptions, and patient-reported barriers to adherence, enabling systematic assessment of treatment non-adherence across large patient populations.

In this study, we examine hypertension treatment non-adherence using EHR data from UCSF by leveraging an LLM to analyze clinical notes, and further investigate its impact on causal inference and ML model performance (Figure~\ref{fig:diagram}). With a cohort of 3,623 patients, we identify 786 (21.7\%) cases of non-adherence and extract demographic and clinical factors that are statistically significant. Additionally, we apply topic modeling to clinical notes revealing underlying reasons for non-adherence.

To assess the effect of treatment non-adherence bias on downstream model performance, we perform causal inference and build predictive models using EHR with treatment records.  Our results show that ignoring treatment non-adherence bias could lead to reversed conclusions in treatment effect estimation, significantly degrade the performance of predictive models up to 5\%, and lead to unfair predictions. Furthermore, we highlight the importance of addressing treatment non-adherence bias by showing simply removing patient records with non-adherence, though reducing the size of the training dataset, could improve model performance and lead to fairer predictions.
\\
\newline
The contributions of this work include:
\begin{enumerate}
\item Conducting a large-scale study on treatment non-adherence in hypertension and identifying statistically significant factors associated with non-adherence.

\item Comparing LLM identification against physician annotations, LLMs perform well with 92\% accuracy, precision and recall.

\item Identifying patient-reported reasons for treatment non-adherence including side effects, forgetfulness, difficulties obtaining refills, etc.

\item Demonstrating the harmful impact of ignoring treatment non-adherence bias on causal inference and predictive modeling, leading to poorer performance and exacerbating racial disparities.

\end{enumerate}

\section{Related Work}
\label{sec:related_work}
\subsection{Treatment adherence analysis in hypertension}
Multiple studies have investigated treatment adherence among patients with hypertension \citep{boratas2018evaluation, uchmanowicz2018factors, algabbani2020treatment, najjuma2020adherence, schober2021high}. These studies are mainly cross-sectional, with a cohort of admitted patients collected at a fixed time point, and treatment adherence is typically measured through questionnaires and interviews. For instance, \citet{algabbani2020treatment} conducted a study in Saudi Arabia involving 306 hypertensive outpatients, finding that only 42.2\% of participants adhered to their antihypertensive medications. \citet{boratas2018evaluation} conducted a similar study of 147 hypertensive patients, identifying factors such as age and duration of hypertension to be significant. However, due to their reliance on questionnaire and interview data, they often have small sample sizes (e.g., less than 300 patients) and self-reporting bias \citep{adams1999evidence, stirratt2015self}, which limits their representativeness and can even lead to contradictory conclusions. In contrast, our work conducts the first large-scale analysis utilizing EHR, with a significantly larger sample size of 3,623 patients.

\subsection{Machine learning and treatment adherence}
Machine learning has been used to identify individual risk factors associated with treatment non-adherence \citep{koesmahargyo2020accuracy, gichuhi2023machine, burgess2023using}. \citet{gichuhi2023machine} developed ML algorithms and found SVM achieved 91.28\% accuracy in predicting tuberculosis treatment non-adherence. Instead of predicting treatment adherence, our work focuses on analyzing the impact of treatment non-adherence bias on downstream model performance. Other studies have applied natural language processing (NLP) to analyze surveys to better understand treatment non-adherence \citep{anglin2021natural, lin2022extraction, chan2024patient}. \citet{chan2024patient} applied NLP to free-text responses from questionnaires completed by type 2 diabetes patients, identifying key reasons for non-adherence. Unlike questionnaires, our work leverages treatment adherence information extracted from clinical notes using LLMs. Lastly, \citet{zhong2022use} applied ML while accounting for adherence information when analyzing treatment effects in a randomized controlled trial. To our knowledge, our study is the first to leverage LLMs for extracting treatment adherence information from clinical notes and evaluating its impact on downstream causal inference and predictive model performance.

\section{Study Design}
\label{sec:setup}

\subsection{Hypertension cohort selection}
We identified 15,002 patients with primary hypertension and extracted their primary care visits occurring on or after January 1st, 2019 following their initial hypertension diagnosis. To assess treatment adherence, consecutive visits for each patient were grouped into pairs. We focused on pairs where a hypertension medical prescription was provided during the first visit, and verified adherence at the second visit by extracting the associated clinical notes. 

Our analysis focuses on ten commonly prescribed hypertension medications: amlodipine, losartan, lisinopril, benazepril, carvedilol, hydralazine, hydrochlorothiazide, clonidine, spironolactone, and metoprolol \citep{heartTypesBlood}. Therefore, we excluded pairs in which the first visit lacked a medication record on this list, as well as pairs with missing or invalid notes during the second visit. We further focus on pairs where the interval between visits is between one month and one year. Lastly, we filtered out patients with unknown demographic information for the purpose of analysis. This resulted in a final cohort of 3,623 patients with 5,952 visit pairs. The cohort selection process is summarized in Appendix~\ref{apd:cohort}.

Demographic information, including sex, age, race, and marital status, was extracted from patient records. Four clinical factors were further derived from the EHR, many of which have been shown to be associated with hypertension non-adherence  \citep{boratas2018evaluation, algabbani2020treatment}. These factors include the duration between the two visits in the pair, the duration of hypertension, the number of primary care visits and the number of comorbidities. We quantified comorbidities using the Charlson Comorbidity Index (CCI) \citep{charlson1987new} and the Elixhauser Comorbidity Index (ECI) \citep{elixhauser1998comorbidity}, which condensed diagnoses into 17 and 31 well-defined comorbidity categories respectively. The demographic and clinical characteristics of the selected cohort are summarized in Table~\ref{tab:medication_adherence}. We detail the comorbidity categories along with other features used in the study in Appendix~\ref{apd:feature}.

\subsection{LLM configuration and prompt engineering}
We used the GPT-4o model \citep{openai2024gpt4ocard} (version 2024-05-13) via the HIPAA-compliant Microsoft Azure API, with the temperature set to 0 and all other parameters left at default. For each pair of visits, we provided the prescription record from the first visit and the clinical notes from the second visit to the model to assess adherence to the prescribed medication.

The model was prompted to identify instances of non-adherence, the type of non-adherence, and extract relevant sections from the notes. We used a zero-shot approach without additional training data or fine-tuning. We also implemented a second round of prompt validation by feeding the model's initial output back into the model, asking it to double-check its response. This additional step significantly reduced hallucinations. The prompt used in the study is provided in Appendix~\ref{apd:prompt}.

The cost for running all GPT-4o evaluations, including prompt development and inference was \$184.77, based on a cost of \$0.005 per 1,000 input tokens and \$0.015 per 1,000 output tokens.

\subsection{Physician validation of LLM detection}

To ensure the reliability of the LLM detection, we randomly selected 50 pairs labeled by the model as non-adherence and 50 pairs labeled as adherence for physician validation to assess accuracy. The gold standard was established through physician annotations conducted independently of the model's predictions. Overall, the model achieved an accuracy of 92\%, with four instances of physician-labeled non-adherence not detected and four adherent instances mislabeled as non-adherence(92\% precision and recall).

We further analyze discrepancies between the model and physician annotations, noting that some mismatches arise from ambiguous notes. For example, cases where patients restarted medication after hospitalization were marked as non-adherent by the LLM, since treatment was paused during hospitalization. Whereas physicians labeled them as adherent, considering the pause as a temporary interruption rather than true non-adherence.

\section{Treatment Non-adherence Analysis}
\label{sec:analysis}
We begin by presenting the results of the identified hypertension treatment non-adherence with statistical testing in Section~\ref{sec:treatment_adherence_analysis}. In Section~\ref{sec:treatment_adherence_clustering}, we apply topic modeling to the extracted clinical notes, uncovering underlying reasons contributing to treatment non-adherence.

\subsection{Factors associated with treatment non-adherence}
\label{sec:treatment_adherence_analysis}
\begin{table*}[t]
\centering
\caption{Demographic and clinical characteristics of patients in the study and logistic regression results. Age, one-hot encoding for Black race, and the number of comorbidities are found to be statistically significant factors of treatment non-adherence.}
\label{tab:medication_adherence}
\resizebox{\textwidth}{!}{%
\begin{tabular}{@{}lcccccc@{}}
\toprule
\textbf{Factors}  & \textbf{Total} & \textbf{Non-adherent} & \textbf{Adherent} & \multicolumn{3}{c}{\textbf{Bivariate Analysis}} \\ 
\cmidrule(l){5-7}
& $n=3623$ & $n=786 (21.7\%)$ & $n=2837 (78.3\%)$ & \textbf{Unadjusted OR} & \textbf{95\% CI} & \textbf{$p$-value} \\ \midrule
\textbf{Demographics}  & & & & & & \\ 
\hspace{1em} Sex & & & & & & \\
\hspace{2em} Female & 2143 & 473 & 1670 & Ref. & Ref. &  \\
\hspace{2em} Male & 1480 & 313 & 1167 & 0.95 & (0.81 to 1.11) & 0.508\\ 
\hspace{1em} Age, mean ± SD & 62.03 ± 14.2 & 61.09 ± 14.7 & 62.29 ± 14.1 & 0.94 & (0.89 to 1.00) & \textbf{0.036} \\
\hspace{1em} Race & & & & & & \\
\hspace{2em} Asian & 1125 & 244 & 881 & Ref. & Ref. &  \\
\hspace{2em} Black & 419 & 114 & 305 & 1.35 & (1.04 to 1.75) &  \textbf{0.023}\\ 
\hspace{2em} White & 1646 & 327 & 1319 & 0.90 & (0.74 to 1.08) &  0.244\\ 
\hspace{2em} Other & 433 & 101 & 332 & 1.10 & (0.84 to 1.43) & 0.486\\
\hspace{1em} Marital Status & & & & & & \\
\hspace{2em} Divorced & 329 & 69 & 260 & Ref. & Ref. &  \\
\hspace{2em} Married & 1861 & 370 & 1491 & 0.94 & (0.70 to 1.25) &  0.649\\ 
\hspace{2em} Single & 878 & 220 & 658 & 1.26 & (0.93 to 1.71) &  0.139\\ 
\hspace{2em} Widowed & 358 & 77 & 281 & 1.03 & (0.72 to 1.49) & 0.864\\
\hspace{2em} Other & 197 & 50 & 147 & 1.28 & (0.85 to 1.94) & 0.243\\
\textbf{Clinical Factors}  & & & & & & \\ 
\hspace{1em} Time between visits (days), mean ± SD & 116.89 ± 83.4 & 112.04 ± 85.5 & 118.23 ± 82.8 & 1.00 & (1.00 to 1.00) & 0.066 \\
\hspace{1em} Total number of comorbidities (ECI), mean ± SD & 3.13 ± 2.4 & 2.98 ± 2.2 & 3.17 ± 2.4 & 0.96 & (0.93 to 1.00) & \textbf{0.045}\\
\hspace{1em} Duration of hypertension (years), mean ± SD & 5.94 ± 6.5 & 5.75 ± 6.6 & 6.00 ± 6.5 & 0.99 & (0.98 to 1.01) & 0.341\\
\hspace{1em} Number of primary care visits one year prior the visit, mean ± SD & 15.75 ± 11.6 & 15.71 ± 10.8 & 15.76 ± 11.8 & 1.00 & (0.99 to 1.01) & 0.925\\
\bottomrule
\end{tabular}
}
\end{table*}

\begin{table}[h]
\centering
\footnotesize 
\caption{Multivariate logistic regression results of factors associated with treatment non-adherence. The number of comorbidities and one-hot encoding for Black race remain statistically significant.}
\label{tab:medication_adherence_multivariable}
\begin{tabular}{@{}l@{\hspace{-12pt}}c@{}c@{\hspace{5pt}}c@{}}
\toprule
\textbf{Factors}  &  \multicolumn{3}{c} \textbf{Multivariate Logistic Regression}\\ 
\cmidrule(l){2-4}
& \textbf{Adjusted OR} & \textbf{95\% CI} & \textbf{$p$-value} \\ 
\midrule
Age (per 10-year increment) & 0.97 & (0.91 to 1.02) & 0.242 \\
Number of comorbidities & 0.96 & (0.93 to 1.00) & \textbf{0.03}\\
 Race & & &\\
\hspace{1em} Asia & Ref. & Ref. &  \\
\hspace{1em} Black & 1.38 & (1.06 to 1.80) &  \textbf{0.016}\\ 
\hspace{1em} White & 0.90 & (0.75 to 1.08) &  0.266\\ 
\hspace{1em} Other & 1.10 & (0.84 to 1.43) & 0.5\\
\bottomrule
\end{tabular}
\end{table}
To meet the independence assumption of the statistical tests, we keep only the most recent pair of visits for adherent patients and the most recent non-adherent pair for non-adherent patients when multiple pairs are available for the same patient.

Among 3,623 patients, 786 (21.7\%) are identified as non-adherent to their treatment plans. Of these 786 patients, 506 (64.4\%) miss their prescribed treatments, 237 (30.2\%) take a different dosage than instructed, 53 (6.7\%) use a different medication, and 62 (7.9\%) take their medication at a time other than instructed. Note that a single patient may exhibit multiple types of non-adherence above.

To identify factors associated with nonadherence to treatment, we begin by performing unadjusted logistic regression, with the results including confidence intervals and p-values presented in Table~\ref{tab:medication_adherence}. 
We find three factors that are statistically significant ($p<0.05$): age ($p=0.036$), one-hot encoding for Black race ($p=0.023$), and the number of comorbidities ($p=0.045$).

Our findings indicate that younger patients are less likely to adhere to treatment, aligning with previous research \citep{boratas2018evaluation} that suggests adherence improves with age as patients become more accustomed to managing their diagnoses. Additionally, we find that Black patients exhibit higher rates of non-adherence. In the U.S., Black individuals have a higher prevalence of uncontrolled hypertension than White individuals \citep{aggarwal2021racial}, and our finding further highlighting the need for greater attention to prevent further exacerbation of racial disparities in hypertension control. Lastly, a lower number of comorbidities is associated with a higher rate of non-adherence, possibly because patients with fewer health conditions may perceive their treatment as less essential.

To avoid potential confounding, we then adjust for the identified significant factors in the multivariate logistic regression model. Results are presented in Table~\ref{tab:medication_adherence_multivariable}, where we see that the number of comorbidities ($p=0.03$) and race as Black ($p=0.016$) still remain statistically significant.

\subsection{Uncovering reasons for treatment non-adherence}
\label{sec:treatment_adherence_clustering}

\begin{figure}[h]
    \centering
    \includegraphics[width=\linewidth]{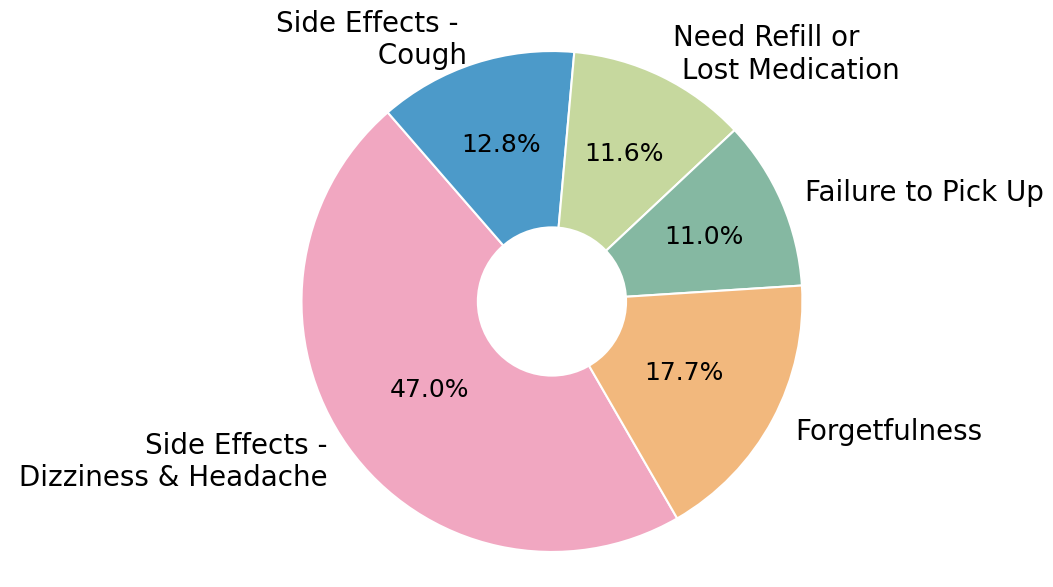}
    \caption{BERT topic modeling results for treatment non-adherence reasons. Side effects are the dominant reason for non-adherence, and 17.7\% of reasons are due to forgetfulness, while others are related to not picking up the medication, needing a refill, or losing it.}
    \label{fig:reason}
\end{figure}

We employ Bidirectional Encoder Representations from Transformers (BERT)-based topic modeling \citep{grootendorst2022bertopicneuraltopicmodeling} on the extracted non-adherent clinical notes to uncover reasons. This utilizes a BERT architecture \citep{devlin2019bertpretrainingdeepbidirectional} to generate embeddings from the extracted notes and applies Uniform Manifold Approximation and Projection (UMAP) \citep{mcinnes2020umapuniformmanifoldapproximation} to reduce the dimensionality of the embeddings. The reduced embeddings are subsequently clustered using HDBSCAN \citep{campello2013density}, and key terms for each cluster are identified using class-based Term Frequency-Inverse Document Frequency (c-TF-IDF).

Across all non-adherent instances, 164 clinical notes fall into clusters. The full clustering with key terms found is in Appendix~\ref{apd:topic}, and we summarize the topics in Figure~\ref{fig:reason}. 59.8\% of the extracted reasons for non-adherence are related to side effects, with most patients experiencing dizziness and headaches after taking the prescribed medication, while a few report coughing. 17.7\% of the extracted reasons indicate non-adherence due to forgetfulness. Additionally, 11.6\% are due to needing a refill or losing their medication and 11\% of the reasons involve patients not picking up their medication.

\section{Impact of Non-Adherence Bias on Downstream Performance}
\label{sec:performance}
In this section, we evaluate the impact of treatment non-adherence bias on downstream model performance. We first define the outcome of interest and demonstrate the effect of non-adherence on the outcome in Section~\ref{sec:setup}. We then analyze its impact on treatment effect estimation for causal inference in Section~\ref{sec:causal_inference} and on ML model performance in Section~\ref{supervised_learning}.

\subsection{Treatment non-adherence leads to worse blood pressure outcomes}
\label{sec:setup}
Blood pressure reduction is the primary outcome when evaluating hypertension medications. To investigate the impact of treatment non-adherence on downstream performances, we begin by extracting pairs of visits where blood pressure measurements are available for both visits. To ensure a sufficient number of encounters for each medication, we focus on the top five most commonly prescribed medications: amlodipine, lisinopril, losartan, hydrochlorothiazide and metoprolol. We only include pairs where the duration between visits is less than six months to minimize the influence of other factors that could affect blood pressure over longer intervals. This results in 1732 pairs of encounters in total with 303 non-adherent pairs.

We begin by showcasing the impact of treatment non-adherence on blood pressure reduction between visits using t-tests. The results presented in Table~\ref{tab:treatment_adherence_outcome} indicate that non-adherence leads to smaller blood pressure reduction, with 1.96 mmHg less systolic reduction ($p=0.011$) and 3.93 mmHg less diastolic reduction ($p=0.001$) compared to adherence.

\begin{table}[h]
\centering
\footnotesize 
\caption{Results of the t-tests assessing the effect of treatment non-adherence on blood pressure reduction. Treatment non-adherence is statistically significant for both systolic and diastolic reduction, with non-adherence leading to smaller reductions in systolic and diastolic blood pressure.}
\label{tab:treatment_adherence_outcome}
\begin{tabular}{@{}l@{}c@{\hspace{5pt}}c@{\hspace{5pt}}c@{}}
\toprule
\textbf{Outcome} &  \textbf{Mean Difference} &  \textbf{95\% CI} &\textbf{$p$-value} \\
\midrule
Systolic Reduction &  -1.96 & (-3.47 to -0.46) &    \textbf{0.011} \\
Diastolic Reduction &  -3.93 & (-6.23 to -1.63) &    \textbf{0.001} \\
\bottomrule
\end{tabular}
\end{table}

\begin{table*}[h]
\centering
\footnotesize 
\caption{Estimated ATE of medication on blood pressure reduction using different causal inference methods. Notably, excluding non-adherent data leads to a lower estimated effect on the diastolic blood pressure reduction and reverses the conclusion on the treatment effect of systolic blood pressure reduction consistently across methods.}
\label{tab:ate_comparison}
\begin{tabular}{@{}lcccccccc@{}}
\toprule
& \multicolumn{4}{c}{\textbf{Diastolic Reduction (mmHg)}} & \multicolumn{4}{c}{\textbf{Systolic Reduction (mmHg)}} \\ 
\cmidrule(lr){2-5} \cmidrule(lr){6-9}
 & IPW & S-Learner & T-Learner & X-Learner & IPW & S-Learner & T-Learner & X-Learner \\ 
\midrule
Full Dataset & 1.75 & 0.77 & 1.44 & 1.51 & -0.06 & -0.05 & -0.12 & -0.14 \\
Adherent Data Only & 1.40 & 0.57 & 0.97 & 0.92 & 0.11 & 0.08 & 0.06 & 0.07 \\
\bottomrule
\end{tabular}
\end{table*}

\subsection{Causal inference for treatment effect estimation}
\label{sec:causal_inference}
Amlodipine and Lisinopril are the most commonly prescribed medications for hypertension, leading to numerous randomized controlled trials (RCTs) comparing their treatment effects \citep{cappuccio1993amlodipine, naidu2000evaluation}. However, due to the high cost of RCTs, various causal inference methods have been developed to estimate treatment effects from observational data \citep{pearl2009causality, austin2011introduction, shalit2017estimatingindividualtreatmenteffect, K_nzel_2019}. Among them, Inverse Probability Weighting (IPW) provides an unbiased estimation of the Average Treatment Effect (ATE) by adjusting for confounding \citep{austin2011introduction}. Meta-learners such as the S-learner, T-learner and X-learner have been proposed to improve treatment effect estimation by leveraging flexible ML models to learn heterogeneous treatment effects \citep{K_nzel_2019}. We start by demonstrating the impact of treatment non-adherence bias on ATE estimation using IPW and meta-learners.
\newline
\textbf{Experiment Setup.} 
We compare the ATE estimation with and without including treatment non-adherent data. Demographic and clinical factors are included as confounders and detailed in Appendix~\ref{apd:feature}. Patients prescribed lisinopril act as the control group, while those taking amlodipine are considered the treated group. The treatment effect is assessed based on the reduction in diastolic and systolic blood pressure between two visits.
\\
\newline
\textbf{Results.} 
The results are presented in Table~\ref{tab:ate_comparison}. Across all models, excluding non-adherent data leads to a lower estimated effect on the diastolic blood pressure reduction and a reversal in the systolic blood pressure reduction. Specifically when using IPW, without filtering for non-adherent data, amlodipine lowers diastolic blood pressure by 1.75 mmHg but increases systolic blood pressure by 0.06 mm Hg compared to lisinopril. After excluding non-adherent data, amlodipine lowers diastolic blood pressure by 1.40 mmHg and also reduces systolic blood pressure by 0.11 mmHg compared to lisinopril. This result shows a reversal in the estimated treatment effect for systolic blood pressure reduction before and after excluding non-adherent data. The same reversal trend also holds across different meta-learners (e.g., using the X-learner on the full dataset, amlodipine reduces systolic blood pressure by 0.14 mmHg compared to lisinopril, whereas after excluding non-adherent data, the X-learner estimates an increase of 0.07 mmHg in systolic blood pressure reduction). The results demonstrate that non-adherence bias can lead to systematic distortions in treatment effect estimation.

\subsection{Supervised learning for treatment outcome prediction}
\label{supervised_learning}
We now demonstrate the impact of treatment non-adherence bias on predictive modeling performance. Following a common setup in the literature \citep{mroz2024predicting, yi2024development}, we use patients' EHR data with treatment prescriptions and blood pressure measurements from their first visit as covariates. The target to predict is whether the blood pressure will be normal at their second visit. Following the guidelines of the American Heart Association \citep{heartUnderstandingBlood}, we define normal blood pressure as having a systolic value of less than 120 and a diastolic value of less than 80. To evaluate model performance in predicting outcomes, we use 500 adherent samples as the test set in all subsequent experiments. We test exclusively on adherent patients since our goal is to evaluate model performance in scenarios where treatments are followed as prescribed, representing the intended clinical use case. A detailed description of the features used is provided in Appendix~\ref{apd:feature}.

\begin{figure*}[t]
    \centering
    \includegraphics[width=0.9\textwidth]{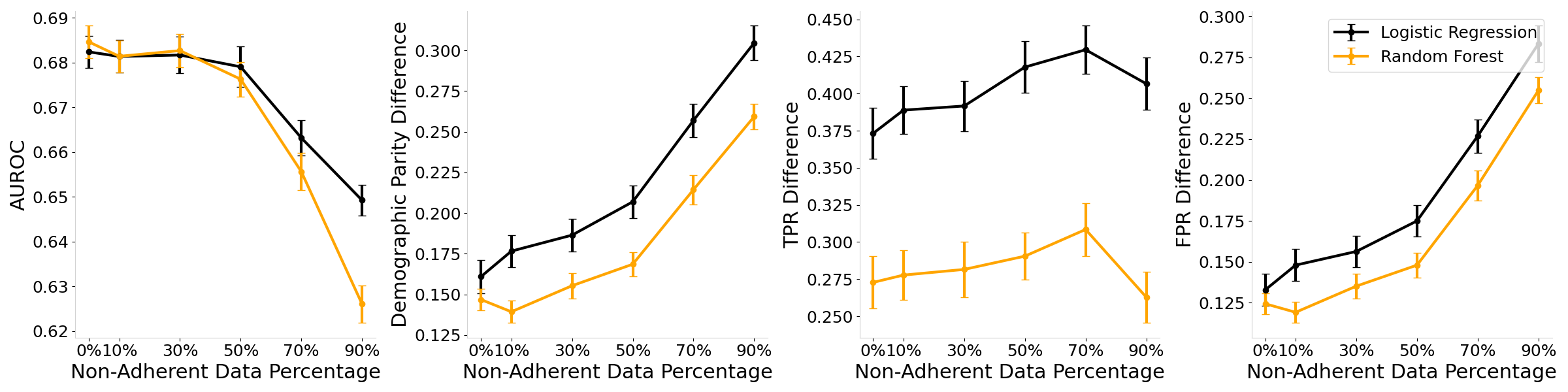}
    \caption{Results of varying treatment non-adherence data percentage on model performance and fairness. Increasing the proportion of non-adherent data in the training set degrades predictive performance and increases fairness disparities between Black and non-Black patients, as measured by demographic parity and the equal odds criterion (true positive rate and false positive rate differences). Results are averaged over 100 seeds, with error bars representing the standard error of the mean.}
    \label{fig:treatment_non_adherence_ratio}
    \includegraphics[width=0.9\textwidth]{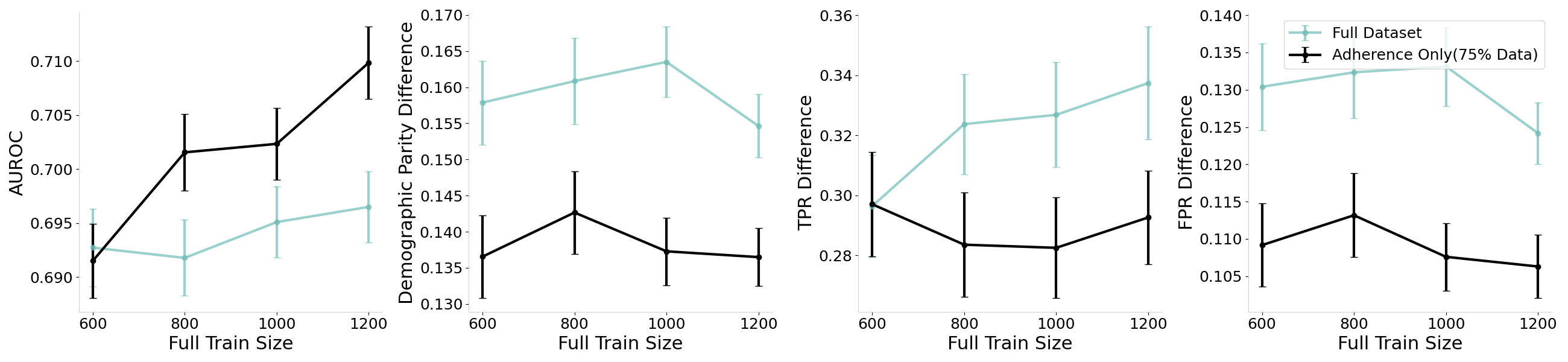}
    \caption{Results of removing non-adherent data on model performance and fairness. The black curve represents training on 75\% of the full dataset, which only consists of adherent encounters. Removing non-adherent data improves model performance, with greater gains observed as sample size increases. It also decreases fairness disparities between Black and non-Black patients, as measured by demographic parity and the equal odds criterion (true positive rate and false positive rate differences). Results are averaged over 100 seeds, with error bars representing the standard error of the mean.}
    \label{fig:remove_non_adherence_rf}
\end{figure*}

\subsubsection{Effect of varying treatment non-adherence data ratios on model performance and fairness}
\label{sec:very_ratio}
We begin by showing the harmful impact of treatment non-adherence bias by varying the proportion of non-adherent data in the training set.
\\
\newline
\textbf{Experiment Setup.}  
We fix the training set size at 300 and vary the proportion of non-adherent data in the training set across \(\{0\%, 10\%, 30\%, 50\%, 70\%, 90\%\}\) to evaluate the impact of treatment non-adherence bias on model performance. We train logistic regression and random forest models, both commonly used for modeling tabular EHR data and assess performance using AUROC on the test set.  Beyond performance, ensuring fair decision-making is also a critical consideration in healthcare. Let \( A \) denote the sensitive attribute (e.g., race), \( Y \) represent the true outcome, and \( \hat{Y} \) denote the predicted outcome. Demographic parity \citep{dwork2012fairness} difference measures the disparity in the likelihood of receiving a positive prediction between groups, i.e.,
\begin{equation*}
    |P(\hat{Y} = 1 \mid A = 1) - P(\hat{Y} = 1 \mid A = 0)|
\end{equation*}
Equal odds \citep{hardt2016equality} difference compares both true positive rates and false positive rates across groups, i.e., 
\begin{align*}
    |P(\hat{Y} = 1 \mid Y = 1, A = 1) - P(\hat{Y} = 1 \mid Y = 1, A = 0)|\\
    |P(\hat{Y} = 1 \mid Y = 0, A = 1) - P(\hat{Y} = 1 \mid Y = 0, A = 0)|
\end{align*}
We therefore measure the differences in demographic parity and equal odds across racial groups to assess fairness.  
\\
\newline
\textbf{Results.} 
We present the results in Figure~\ref{fig:treatment_non_adherence_ratio}, showing that increasing the percentage of non-adherent data in the training set degrades performance, with a 3\% drop for logistic regression and a 5\% drop for random forest in AUROC. Additionally, a higher proportion of non-adherent data increases fairness disparities between Black and non-Black patients under both the demographic parity and equal odds criteria. For instance, the false positive rate disparity of the random forest doubles, increasing from 0.125 to 0.25 as the percentage of non-adherent data increases. Similar trends are observed for other racial groups, and we provide full results in Appendix~\ref{apd:fairness}. These findings consistently highlight the harmful impact of treatment non-adherence bias.

\subsubsection{
Effect of removing non-adherent data on model performance and fairness}
\label{sec:drop_data}
We now emphasize the importance of addressing treatment non-adherence bias by showing that a simple approach to remove non-adherent data can improve predictive performance and lead to fairer predictions.
\\
\newline
\textbf{Experiment setup.}
We fix the non-adherent data ratio at 25\% and compare the performance of random forests trained on the entire dataset versus those trained only on adherent data by removing a quarter of data that are non-adherent. We report the test AUROC as well as demographic and equal odds differences while varying the full training set size in \(\{600, 800, 1000, 1200\}\) before removing non-adherent data.
\\
\newline
\textbf{Results.} 
The results are presented in Figure~\ref{fig:remove_non_adherence_rf}. While the traditional ML perspective suggests that more data generally improves performance, our findings show that using only the adherent 75\% of the data leads to better model performance, with the improvement becoming more significant as the sample size increases. For instance, with a training size of 1,200, the model achieves an AUROC of 0.695 when using all data, whereas dropping non-adherent data improves AUROC to 0.71. Additionally, we find removing non-adherent data reduces racial disparities between Black and non-Black patients, as both demographic parity and equal odds differences are consistently smaller across sample sizes. Similar trends are observed for other racial groups, and we provide full results in Appendix~\ref{apd:fairness}. These findings further highlight the importance of addressing treatment non-adherence bias to achieve better and fairer model performance.

\section{Discussion}
\label{sec:discussion}
Treatment non-adherence is a crucial factor in building treatment models but is often overlooked in practice. By leveraging LLMs with clinical notes, we identify non-adherent encounters with hypertension patients and further demonstrate how treatment non-adherence biases can degrade downstream model performance while exacerbating fairness gaps.

While our study focuses on hypertension, the same pipeline can be applied to other disease areas to analyze treatment non-adherence patterns. Beyond adherence, future work can also utilize LLMs to extract insights from clinical notes on other factors such as medication tolerance, side effects, social determinants of health, and patient-provider communications~\citep{guevara2024large,robitschek2024large,zink2024access,antoniak2024nlp,miao2024identifying}. Our results show that removing non-adherent data from the training set improves both model performance and fairness. Instead of excluding non-adherent data entirely which could lead to insufficient samples, future research could also explore strategies to better integrate non-adherent data into modeling or develop models that are more robust to treatment adherence biases. Additionally, understanding the degree of non-adherence remains a significant challenge due to the absence of standardized quantitative metrics in clinical notes. Future work could explore approaches to estimate non-adherence severity, which may offer deeper insights into non-adherence patterns and enhance more effective strategies for downstream modeling.

Although leveraging LLMs with clinical notes enables large-scale analysis of treatment non-adherence, our study holds keys limitations. First, our work relies on the premise that non-adherence is explicitly documented in the notes, which means cases not mentioned may be missed, potentially leading to an underestimation of the true non-adherence rate. Furthermore, while physician validation confirms that the LLM's output is largely accurate, the use of ML models such as LLMs in shifting and censored patient populations may yield changing performances such that the automated extraction should be scrutinized~\citep{pollard2019turning,yuan2023revisiting,finlayson2021clinician,chen2022clustering}. In conclusion, our work demonstrates the impact of treatment non-adherence bias in predictive modeling and causal inference through a real-world study on hypertension medications. We hope this study raises awareness of treatment non-adherence bias for future research on clinical machine learning.

\bibliography{references}

\appendix
\onecolumn

\section{Cohort Creation} \label{apd:cohort}
\begin{figure}[h]
    \centering
    \includegraphics[width=0.45\linewidth]{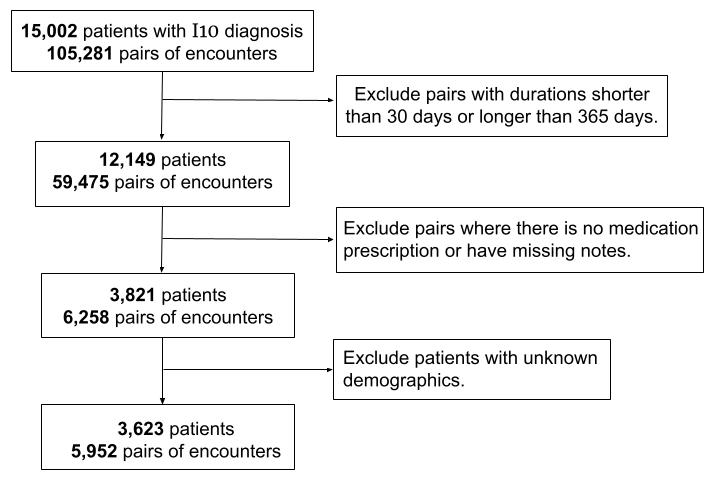}
    \caption{Cohort Selection}
    \label{fig:cohort}
\end{figure}

We detail the cohort selection in Figure~\ref{fig:cohort}.

\section{Feature Definitions}\label{apd:feature}
We use the following features throughout the study. The time between visits is excluded from causal inference and predictive modeling in Section~\ref{sec:performance}, as we instead restrict the analysis to a smaller time window. ECI and CCI are derived using medcodes package \citep{githubGitHubTopspinjmedcodes}.

\begin{table*}[h!]
    \centering
    \small
    \renewcommand{\arraystretch}{1.2} %
    \begin{tabular}{p{4cm} p{12cm}}
        \hline
        \textbf{Feature Name} & \textbf{Description} \\
        \hline
        Sex & The sex of the patients. \\
        Age & The age of the patients at the time of the visit. \\
        Race & The race of the patients. \\
        Marital Status & The marital status of the patients at the time of the visit. \\
        Time Between Visits & The time in days between the two visits in each pair. \\
        Charlson Comorbidity Index (CCI) & A measure of comorbidity based on the following conditions: myocardial infarction, congestive heart failure, peripheral vascular disease, cerebrovascular disease, dementia, chronic pulmonary disease, rheumatic disease, peptic ulcer disease, mild liver disease, diabetes without chronic complications, diabetes with chronic complications, hemiplegia/paraplegia, renal disease, any malignancy, moderate/severe liver disease, metastatic solid tumor, and AIDS/HIV. \\
        Elixhauser Comorbidity Index (ECI) & A measure of comorbidity based on the following conditions: cardiac arrhythmias, congestive heart failure, valvular disease, pulmonary circulation disorders, peripheral vascular disorders, hypertension (uncomplicated or complicated), paralysis, other neurological disorders, chronic pulmonary disease, diabetes (uncomplicated or complicated), hypothyroidism, renal failure, liver disease, peptic ulcer disease, AIDS/HIV, lymphoma, metastatic cancer, solid tumor without metastasis, rheumatoid arthritis, coagulopathy, obesity, weight loss, fluid and electrolyte disorders, blood loss anemia, deficiency anemia, alcohol abuse, drug abuse, psychoses, and depression. \\
        Duration of Hypertension & The duration (in years) between the onset of hypertension and the time of the visit. \\
        Number of Primary Care Visits & The number of primary care visits one year prior to the visit. \\
        \hline
    \end{tabular}
    \caption{Descriptions of features included in the study}
    \label{tab:features}
\end{table*}

\section{Prompt Description}\label{apd:prompt}
\begin{figure*}[h!]
    \centering
    \includegraphics[width=0.95\textwidth]{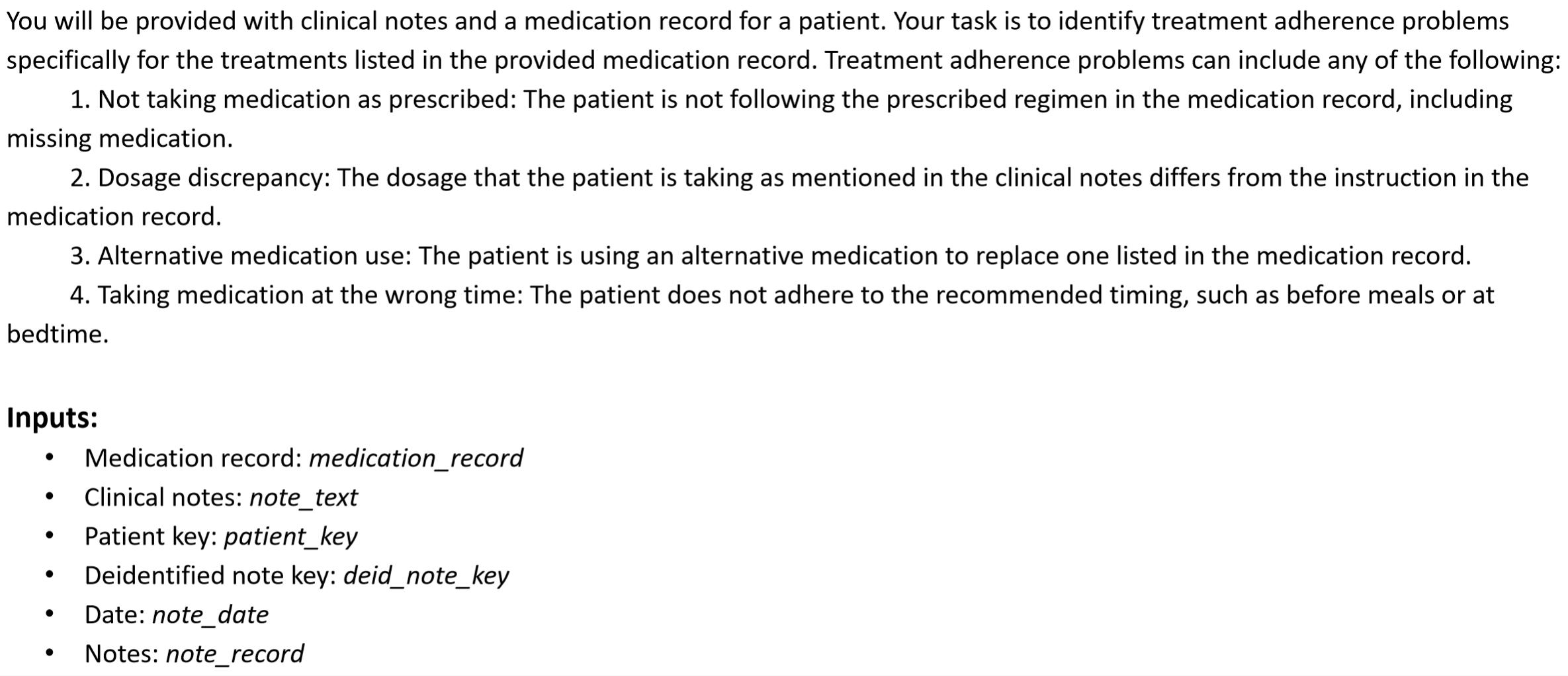} \\
    \includegraphics[width=0.95\textwidth]{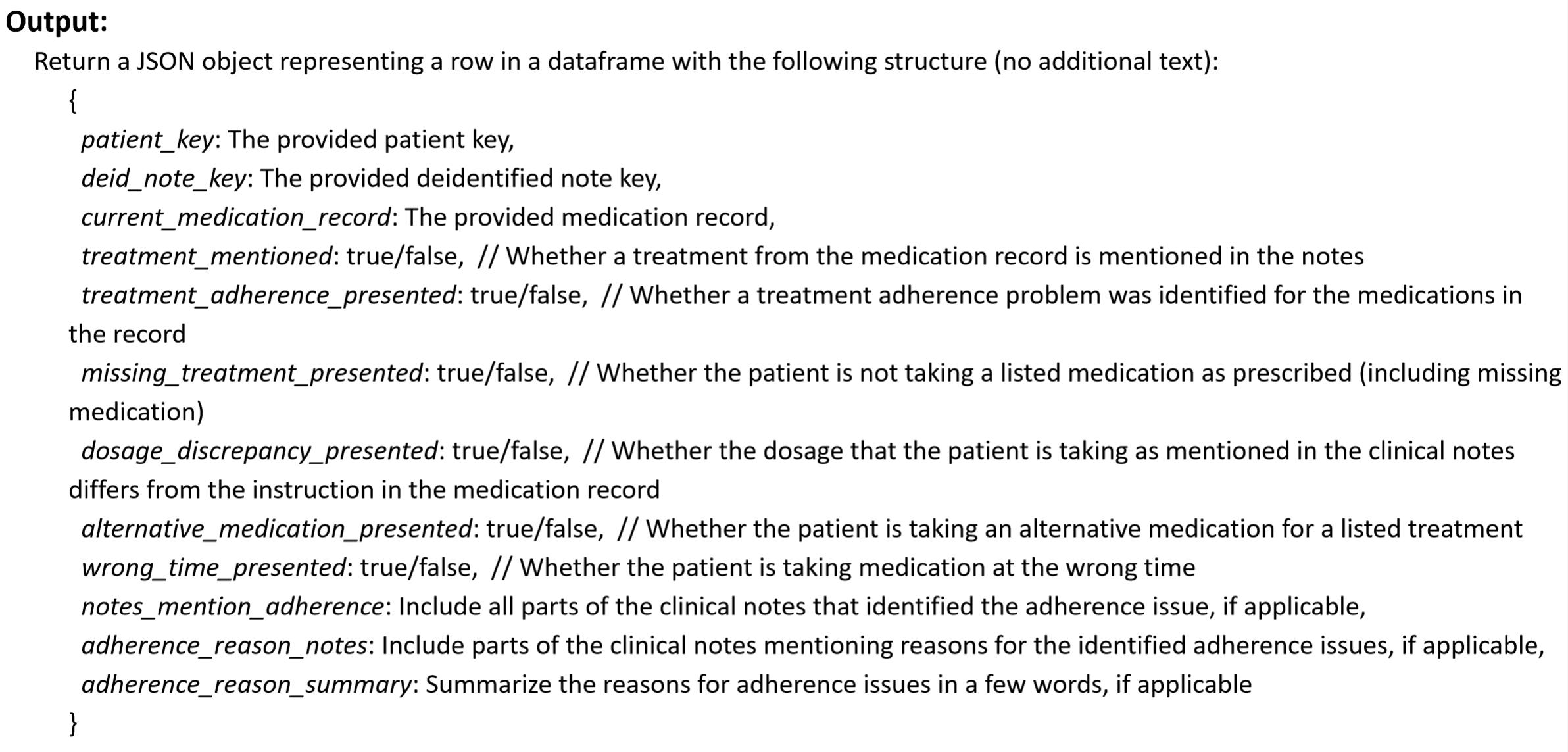} \\
    \includegraphics[width=0.95\textwidth]{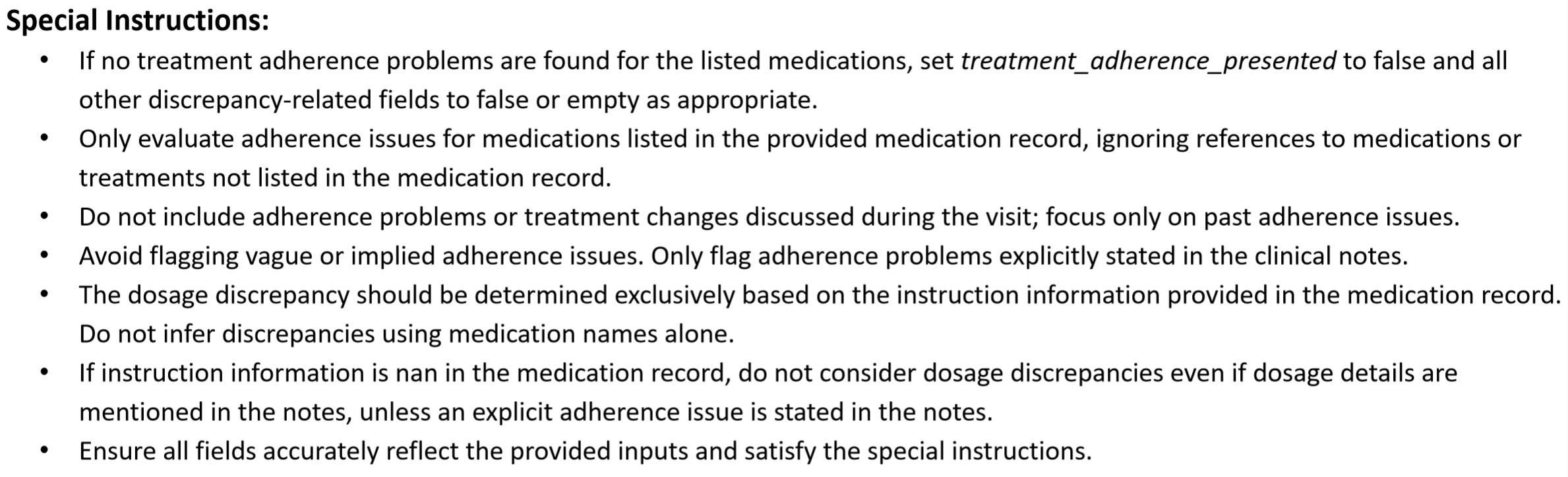}
    \caption{Prompt used in the study}
    \label{fig:prompt}
\end{figure*}

The prompt used in the study is given in Figure~\ref{fig:prompt}.

\section{Topic Modeling}\label{apd:topic}
Here we show the full BERT topic modeling results with key terms found in clusters and we summarize the topics in Figure~\ref{fig:reason}.

\begin{figure*}[h!]
    \centering
    \includegraphics[width=\textwidth]{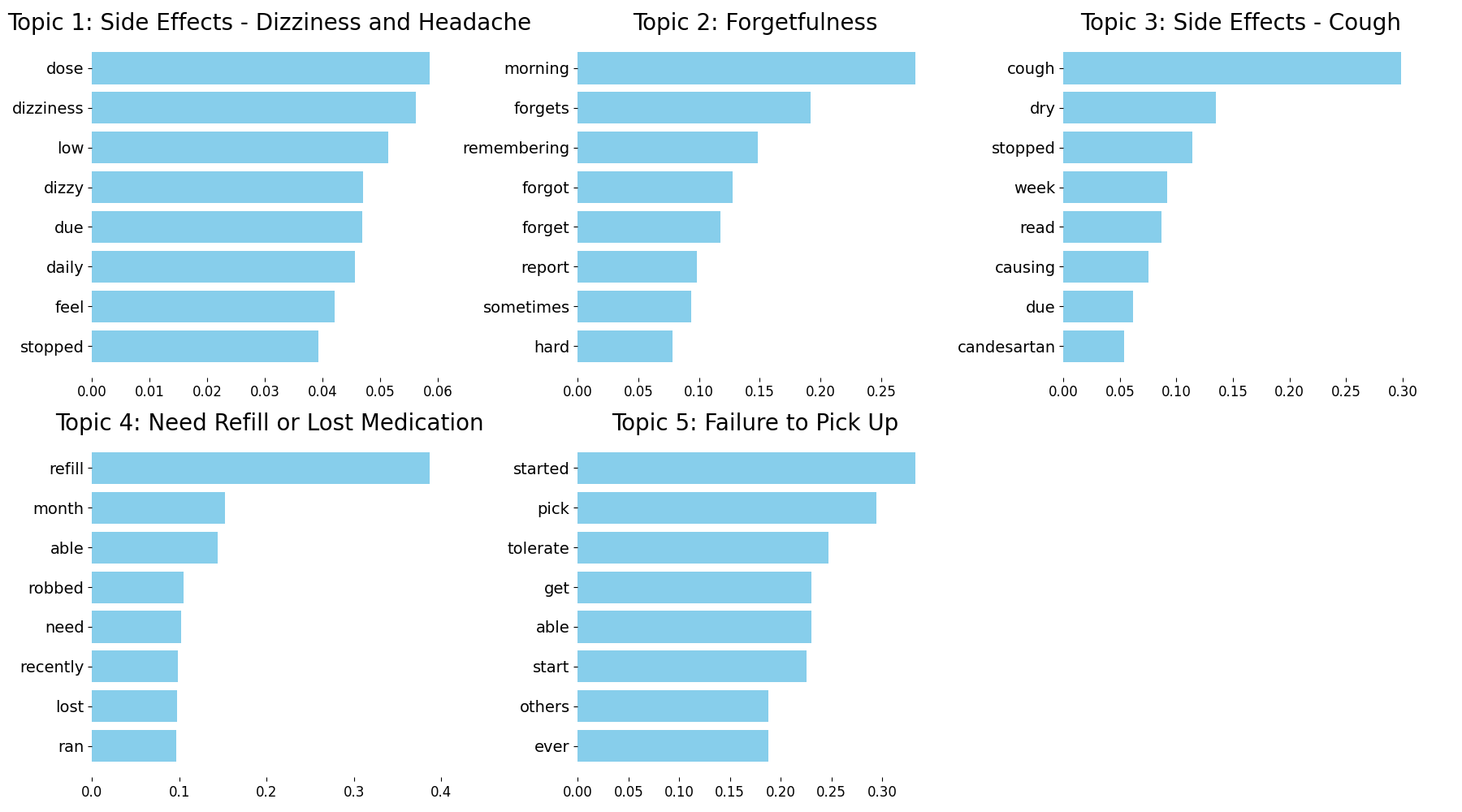}
    \caption{BERT topic modeling identified five clusters of treatment non-adherence reasons. The most common words in each cluster are highlighted in the plot. Key topics identified include side effects, forgetfulness, failure to pick up medications, need for refills, and lost medications. When applying BERT topic modeling, we set a minimum cluster size of 15 notes and used UMAP with 5 components and 15 neighbors for dimensionality reduction.}
    \label{fig:appendix_topic}
\end{figure*}

\newpage
\section{Additional Results of Non-adherence Bias on Predictive Modeling}\label{apd:fairness}
Here, we present the complete results on racial disparities for the experiments in Section~\ref{supervised_learning}.
\begin{figure*}[h!]
    \centering
    \includegraphics[width=\textwidth]{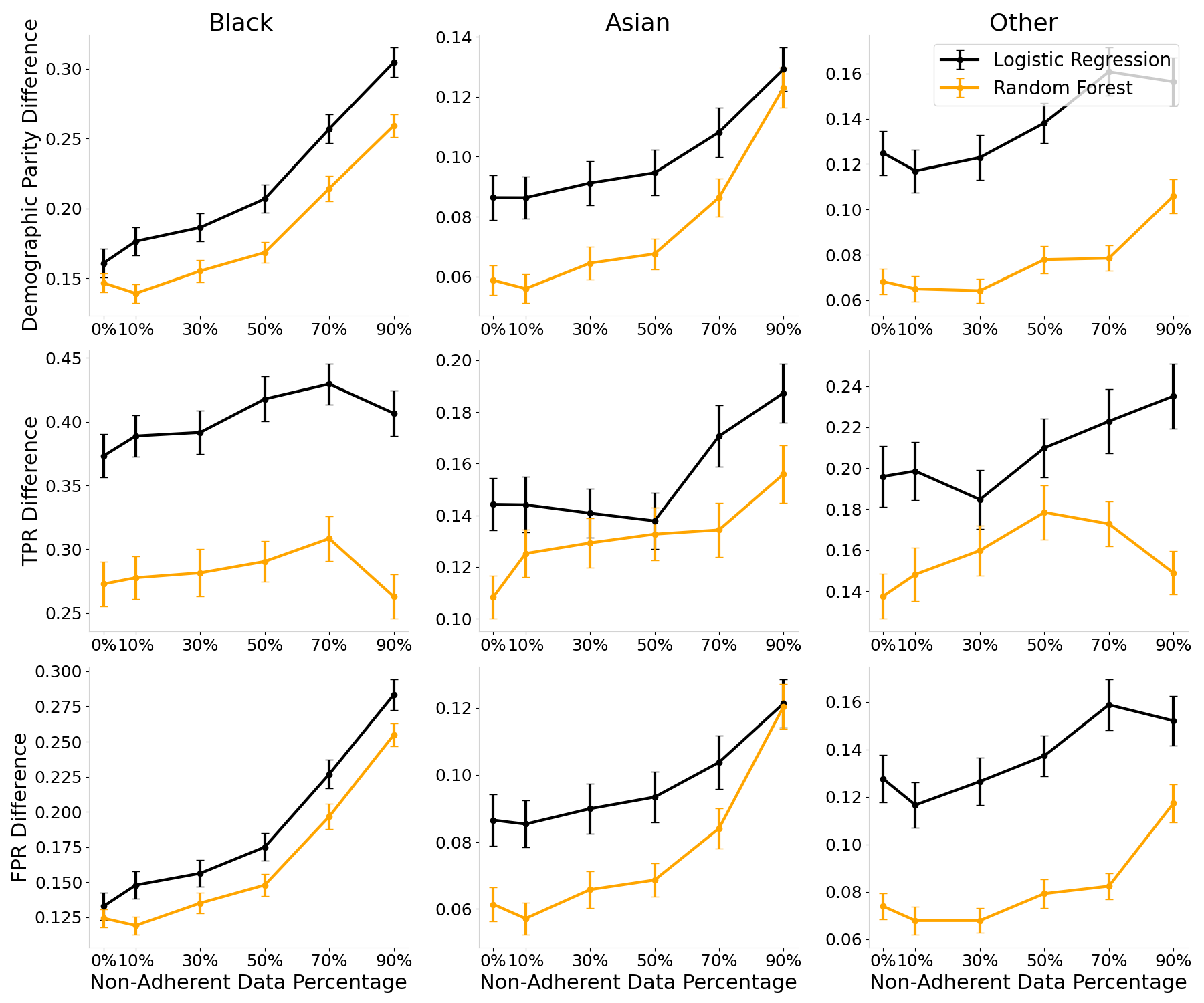}
    \caption{Increasing the proportion of treatment non-adherent data in the training set increases the fairness disparity between different races as measured by demographic parity and the equal odds criterion. Results are averaged over 100 seeds, varying the sampling of the train and test sets. Error bars represent the standard error of the mean.}
    \label{fig:appendix_race_vary_na_ratio}
\end{figure*}
\newpage
\begin{figure*}[h!]
    \centering
    \includegraphics[width=\textwidth]{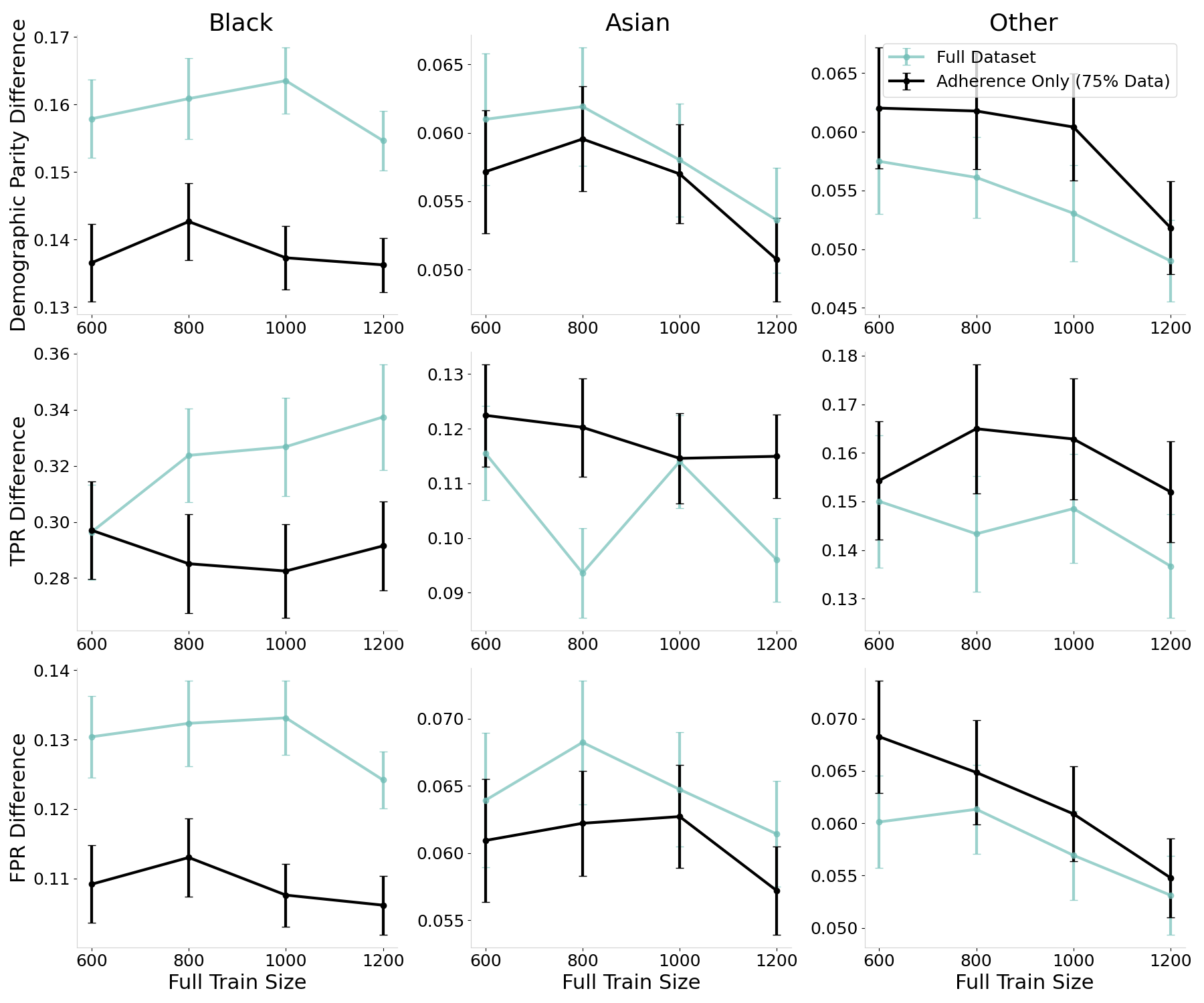}
    \caption{The black curve represents training on 75\% of the full dataset consisting of adherent encounters only. Removing treatment non-adherent data from the training set decreases fairness disparities as measured by demographic parity and the equal odds criterion particularly for Black and Asian. Results are averaged over 100 seeds, varying the sampling of the train and test sets. Error bars represent the standard error of the mean.}
    \label{fig:appendix_race_vary_sample_size}
\end{figure*}

\end{document}